%% file: 00-main.tex
\pdfoutput=1

\documentclass[11pt]{article}

\usepackage[]{acl2021}

\usepackage{times}
\usepackage{latexsym}
\usepackage{graphicx}
\usepackage[T1]{fontenc}
\usepackage{multirow}
\usepackage{booktabs}
\usepackage{array}
\usepackage{graphbox}
\usepackage{bm}
\usepackage{enumitem}
\usepackage{amsmath}
\usepackage{hyperref}
\usepackage{dblfloatfix}

\usepackage[T1]{fontenc}

\usepackage[utf8]{inputenc}

\usepackage{microtype}

\usepackage{textcomp}  
\usepackage{scalerel}  
\def\cosmosemoji{\scalerel*{\includegraphics{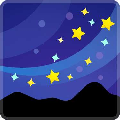}}{\textrm{\textbigcircle}}}
\def\t2cc{COIN}
\def\ourmodel{COSMic}
\def\cc{RaCCoon}

\title{\cosmosemoji~{\fontfamily{qcr}\selectfont COSMic}: A Coherence-Aware Generation Metric for Image Descriptions}

\author{{ Mert İnan} \\ {\normalsize University of Pittsburgh}
        \\ {\small \texttt{mert.inan@pitt.edu}}
        \And
        { Piyush Sharma} \\ {\normalsize Google Research}
        \\ {\small \texttt{piyushsharma@google.com}}
        \And
        { Baber Khalid} \\ {\normalsize Rutgers University}
        \\ {\small \texttt{baber.khalid@rutgers.edu}}
        \AND
        { Radu Soricut} \\ {\normalsize Google Research}
        \\ {\small \texttt{rsoricut@google.com}}
        \And
        { Matthew Stone} \\ {\normalsize Rutgers University}
        \\ {\small \texttt{mdstone@cs.rutgers.edu}}
        \And
        { Malihe Alikhani} \\ {\normalsize University of Pittsburgh}
         \\ {\small \texttt{malihe@pitt.edu}}
        }
        
\begin{document}

\maketitle

\begin{abstract}
Developers of text generation models rely on automated evaluation metrics as a stand-in for slow and expensive manual evaluations. However, image captioning metrics have struggled to give accurate learned estimates of the semantic and pragmatic success of output text. We address this weakness by introducing the first discourse-aware learned generation metric for evaluating image descriptions.  Our approach is inspired by computational theories of discourse for capturing information goals using coherence.  We present a dataset of image--description pairs annotated with coherence relations. We then train a coherence-aware metric on a subset of the Conceptual Captions dataset and measure its effectiveness---its ability to predict human ratings of output captions---on a test set composed of out-of-domain images.
We demonstrate a higher Kendall Correlation Coefficient for our proposed metric with the human judgments for the results of a number of state-of-the-art coherence-aware caption generation models when compared to several other metrics including recently proposed learned metrics such as BLEURT and BERTScore.
\end{abstract}

\input{01-intro}
\input{02-related}
\input{model_figure}
\input{03-data}
\input{04-model}
\input{05-experiments}
\input{06-results}
\input{07-conclusion}

\bibliography{anthology,custom}
\bibliographystyle{acl_natbib}



\end{document}

%% file: 01-intro.tex
\section{Introduction}

\begin{figure}[!ht]
    \centering
    \small
    \begin{tabular}{cp{1.8cm}|c|p{0.6cm}c}
        \multicolumn{5}{c}{\includegraphics[scale=0.06]{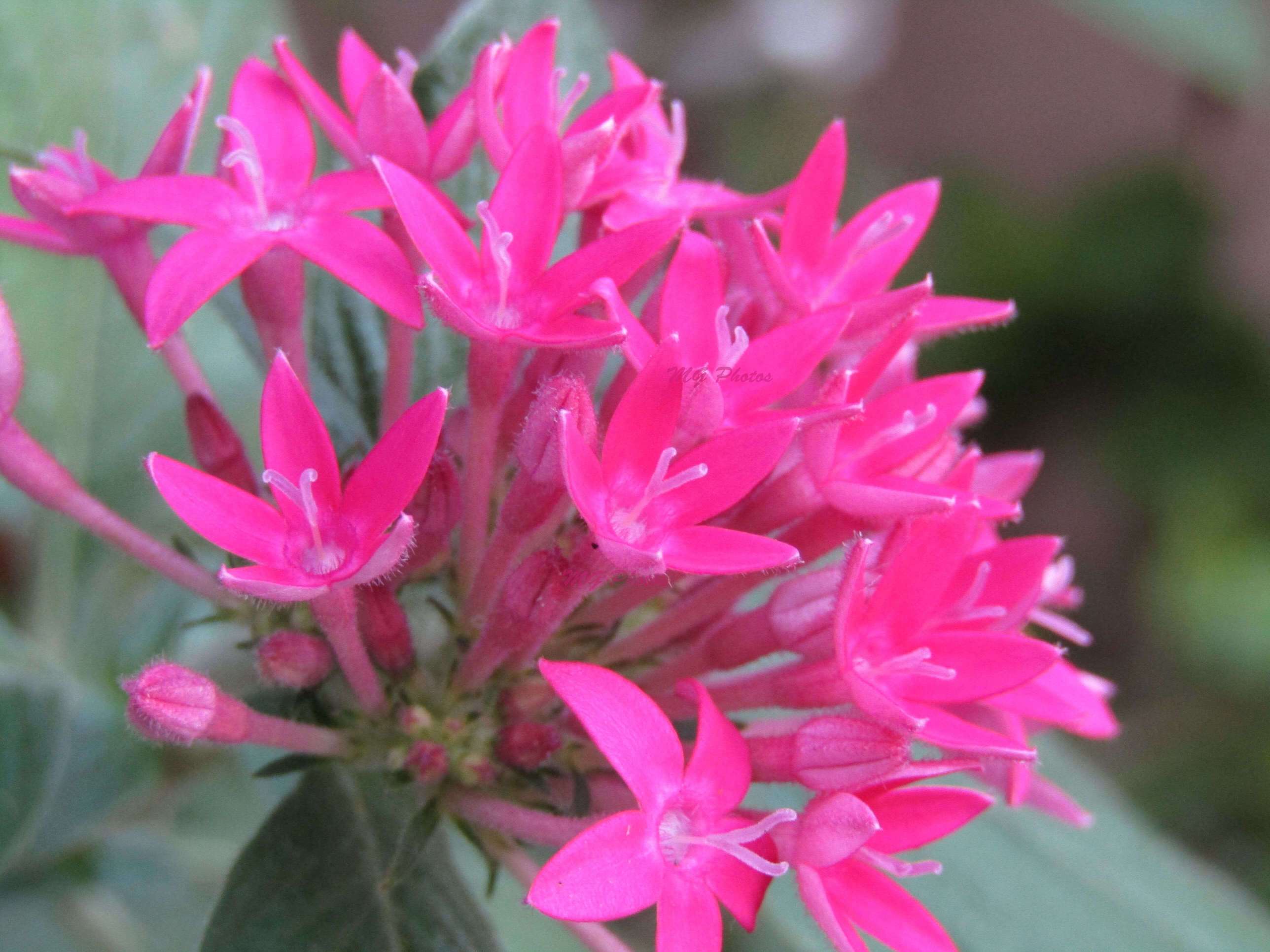}} \\
        \toprule 
        \multicolumn{2}{c|}{Caption} & Coh. & CIDEr & \ourmodel \\
        \midrule
        \multicolumn{1}{c|}{\multirow{2}{*}{Model}} & first flower of the year & \multirow{2}{*}{Story} & \multicolumn{1}{c}{\multirow{4}{*}{0.000}} & \multirow{4}{*}{0.653} \\
        \cmidrule{1-3}
         \multicolumn{1}{c|}{\multirow{2}{*}{Human}} & close-up of pink flowers & \multirow{2}{*}{Visible} & & \\
    \bottomrule
    \end{tabular}
    \caption{
    A comparison of the scores for a generated (Model) caption that has a different coherence relation than the reference (Human) caption. 
    ``Coh.'' represents the coherence labels for generated and reference captions.  
    Our coherence-aware metric \ourmodel\ is aware of the different information goals for these captions, and assigns a more adequate score when comparing the Model caption against the Human caption. In this case where a caption that does not just describe the image but elaborates on it, our metric recognizes that the model output is potentially successful (Photo credit: Moorthy Gounder)}
    \label{fig:intro}
\end{figure}

An investigation of the descriptions used with images on the web shows that image descriptions can have different functions and goals \cite{kruk-etal-2019-integrating,alikhani-etal-2020-cross}. For instance, captions may describe visible entities, activities and relationships, provide background information that goes beyond what's visible, or report the writer's own subjective reactions to what's displayed.
By drawing on such diverse examples, image captioning models can learn the different inferential links between text and images and use that information at  generation time to produce descriptions that can fulfill different discourse goals and inject the desired context into their output \cite{Papineni02bleu:a, lin-2004-rouge, denkowski:lavie:meteor-wmt:2014, DBLP:journals/corr/AndersonFJG16}. 

So far, however, efforts to develop such expressive captioning models have been hindered by the lack of automatic metrics that can evaluate their output with respect to their information goals in context.  
Previous approaches to automatic caption evaluation have mostly focused on n-gram measures of similarity to reference output \cite{DBLP:journals/corr/VedantamZP14a}; such surface-level models fail to deal with the lexical and syntactic diversity of image descriptions. More recent approaches more closely approximate semantic similarity using word embedding-based techniques. These models show robust performance and achieve a higher correlation with human judgments than that of previous metrics.  Nevertheless, they too fail to generalize to the different kinds of content that successful descriptions may exhibit across different goals and contexts. 
%
That is, they cannot distinguish reasonable descriptions that happen to differ from reference output in their goals and perspective, from problematic descriptions that hallucinate inappropriate content or context.

To bridge this gap, we present a coherence-aware embedding-based generation metric that learns to respect diverse discourse goals without penalizing captions that are purposefully generated to fulfill different purposes or communicate background information. Figure \ref{fig:intro} demonstrates this capability by presenting an example image and captions with different coherence labels together with their scores.

Our approach to modeling discourse goals is based on the framework of discourse coherence theory \cite{hobbs-1985}, which characterizes the inferences that give discourse units a coherent joint interpretation using a constrained inventory of coherence relations.  In particular, we use the taxonomy for image--text coherence developed by \newcite{alikhani-etal-2020-cross}, which for example includes \textit{Visible}, \textit{Story} and \textit{Subjective} relations between the text and the image. A description and an image stand in a \textit{Visible} relation if the text includes information that is recognizably depicted in the image. \textit{Subjective} captions  react to the content of the image and \textit{Story} captions provide a free-standing description of the circumstances depicted in the image similar to the \textit{Narration} relation in text. 
Our metric is learned in part from a new dataset of 4000 images with descriptions labeled with different coherence labels in this taxonomy. 


In inaugurating the study of coherence-aware generation metrics, we make the following specific contributions.  In Section~\ref{sec:data} we present two different, annotated datasets for training and testing a coherence-aware metric. We present a model to score a generated caption given the image, reference caption, and the discourse goals of both these captions (Section~\ref{sec:models}).
We compare this metric to previous ones using a common methodology, ranking the performance of several different caption generation systems on out-of-domain images---relying on a new benchmark out-of-domain test set, which we publish, providing reference captions for a subset of OpenImages \cite{kuznetsova2020open}.
Our experiments demonstrate that among all these metrics, our proposed metric has the highest correlation with human judgments.

%% file: 02-related.tex
\section{Related work}

There are diverse ways of characterizing the contributions of text and imagery. \newcite{gao2015you} investigate the genre of image captions and \newcite{huang2016inferring} study the persuasive implicit relationships between text and images. \newcite{kruk2019integrating} study the emotional links between text and images. \newcite{otto2019understanding} present an annotated dataset of text and imagery that compares the information load in text and images.  However, we build on works that study information-level inferences between discourse units in different modalities such as comic book panels \cite{mccloud1993understanding}, movie plots \cite{cumming2017conventions}, and diagrammatic elements \cite{DBLP:journals/lre/HiippalaAHKLOTS21}. In particular, we use \newcite{alikhani-etal-2020-cross}'s relations that characterize inferences between text and images. 

Coherence-aware models have benefited several NLP tasks such as  gesture interpretation \cite{lascarides2009formal, pustejovsky2020situated}, text summarization \cite{xu2019discourse}, machine comprehension \cite{gao-etal-2020-discern}. The majority of these works use Rhetorical Structure Theory (RST) \cite{mann1987rhetorical} and Penn Discourse TreeBank (PDTB) \cite{prasad2008penn} datasets to learn and predict these relations between two adjacent text spans. 
In this line of work, we are the first to present a coherence-aware generation metric. 

\input{t2_gt_images}

The most widely used automatic evaluation metrics are ngram-based, which compute the exact number of ngram matches between reference and generated text \cite{cui2018learning}. Examples of such metrics that are commonly used for evaluating the output of captioning, translation and summarization models are BLEU \cite{Papineni02bleu:a}, ROUGE \cite{lin-2004-rouge}, and CIDEr \cite{vedantam2015cider}, . 
The major problem of the n-gram similarity 
metrics is that they give no credit to synonym 
matches of reference n-grams, even if those words are common and used 
appropriately in the generated text.  Embedding-based metrics such as BLEURT \cite{sellam-etal-2020-bleurt} and BERTScore  \cite{bert-score} designed to address this limitation are closer to human ratings. BLEURT is a data-intensive training scheme that is based on BERT \cite{devlin-etal-2019-bert} fine-tuned on human ratings of generated text. BERTScore, however, computes the similarity score as the average of cosine similarities between predicted tokens and their top matching reference tokens. 
These metrics however, do not respect the information goal and the purpose for which the model has generated the text. We address this problem by introducing the first coherence-aware generation metric. Similar to SPICE \cite{anderson2016spice} and VIFIDEL \cite{madhyastha-etal-2019-vifidel} we use the information encoded in images. We further propose the addition of coherence relations that facilitate learning with fewer samples by a multimodal metric using pre-trained BERT and ViLBERT. 

%% file: t2_gt_images.tex
\begin{figure*}[!ht]
    \centering
    \small
    \begin{tabular}{p{3.6cm}p{3.6cm}p{3.6cm}p{3.6cm}}
        \multicolumn{1}{c}{\includegraphics[height=2.2cm,width=3.6cm]{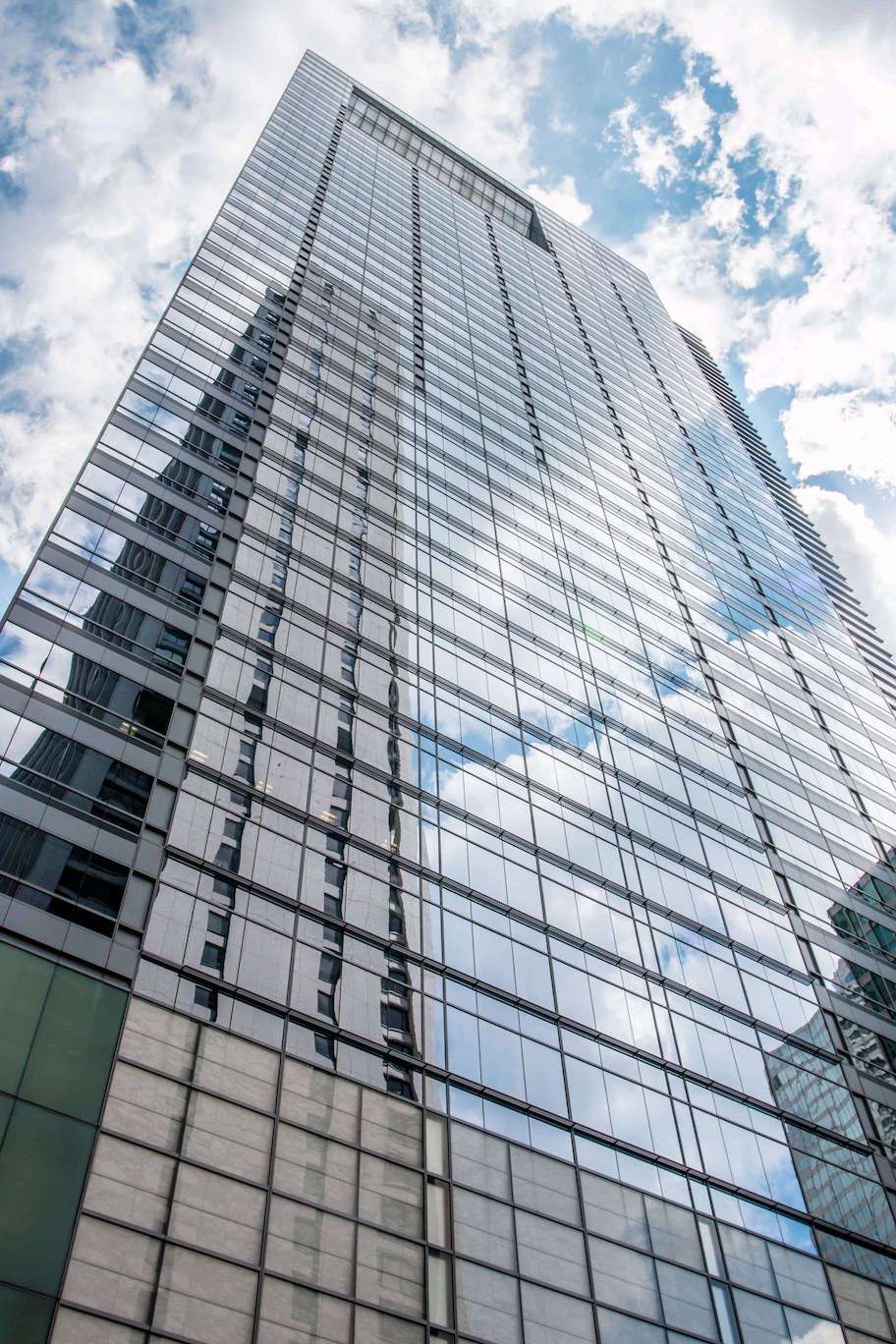}} & \multicolumn{1}{c}{\includegraphics[height=2.2cm,width=3.6cm]{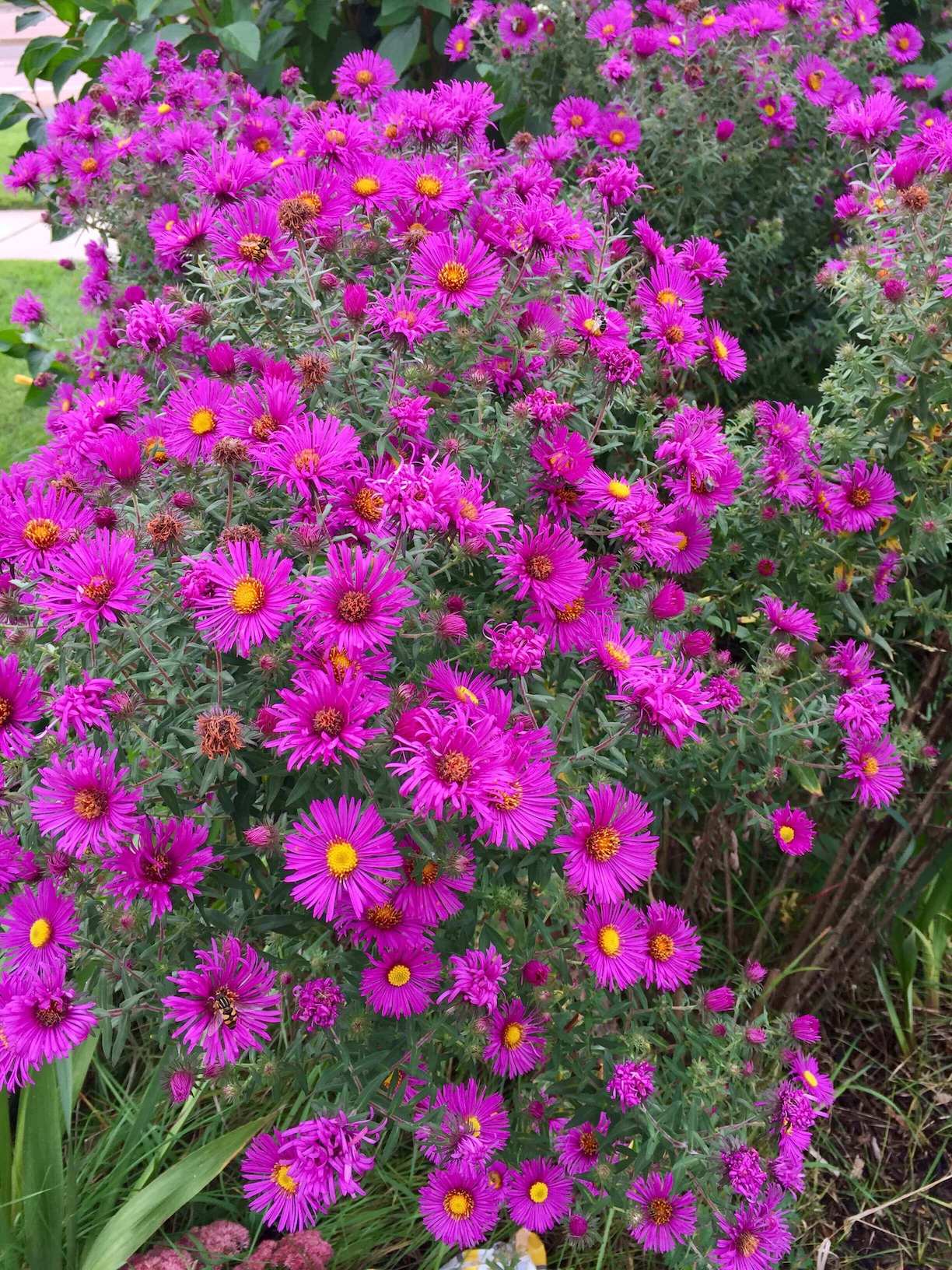}} & \multicolumn{1}{c}{\includegraphics[height=2.2cm,width=3.6cm]{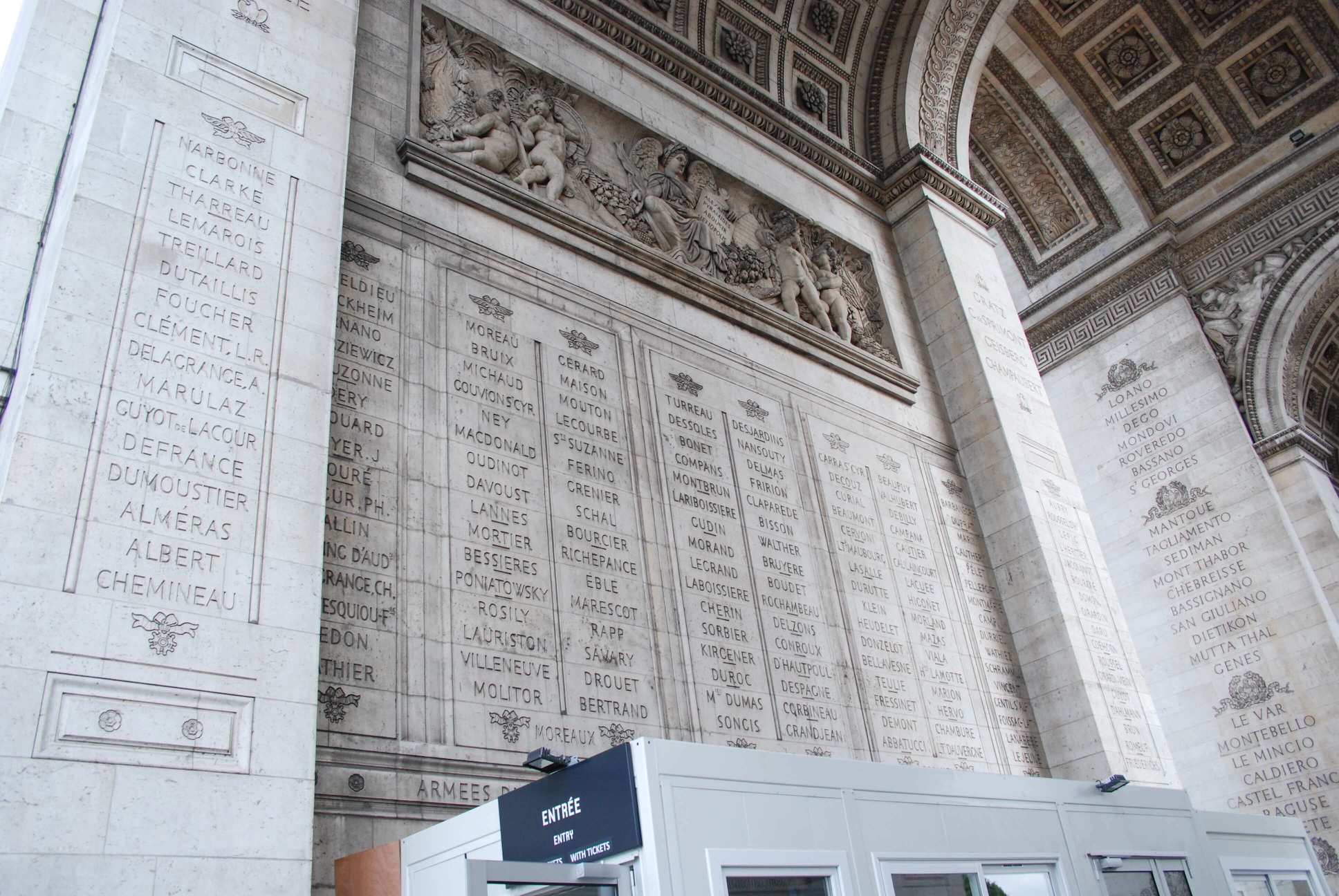}} & \multicolumn{1}{c}{\includegraphics[height=2.2cm,width=3.6cm]{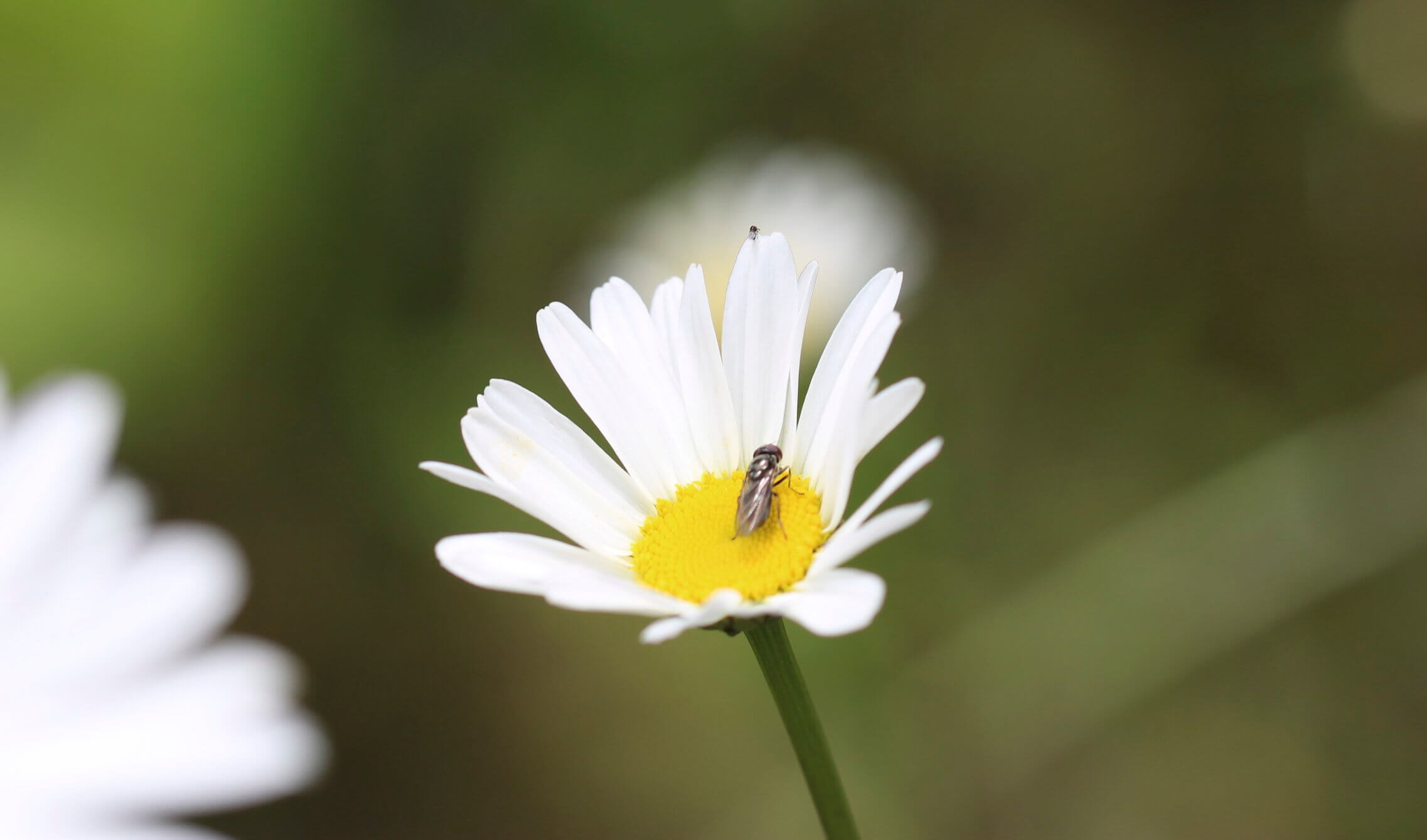}}\\
         Facade of a glass building. &  A pink flower bush in a garden. &  The underside of the Arc de Triomphe. &  Close-up of a fly sitting on a daisy. 
         \\
        \multicolumn{1}{c}{\includegraphics[height=2.2cm,width=3.6cm]{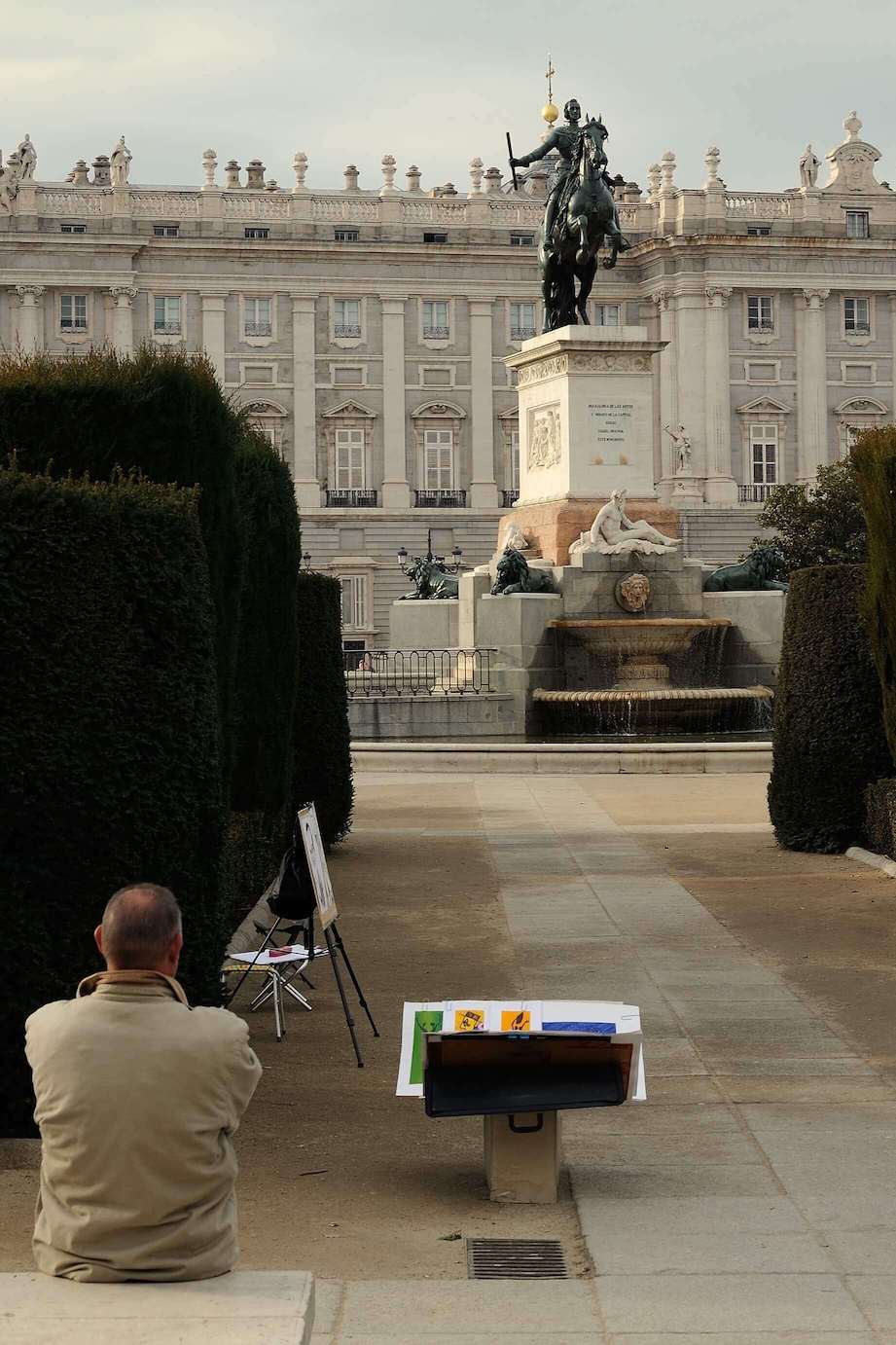}} & \multicolumn{1}{c}{\includegraphics[height=2.2cm,width=3.6cm]{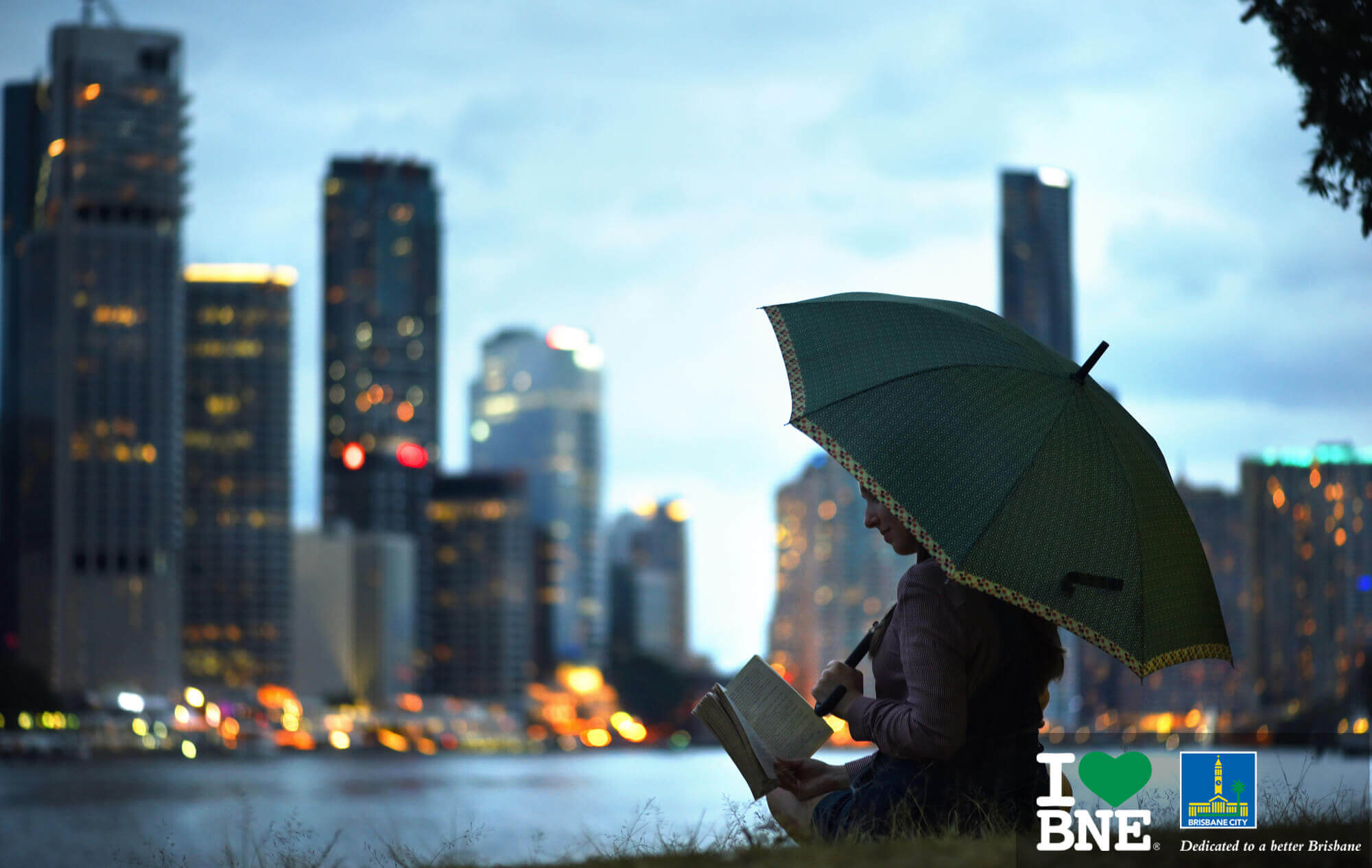}} & \multicolumn{1}{c}{\includegraphics[height=2.2cm,width=3.6cm]{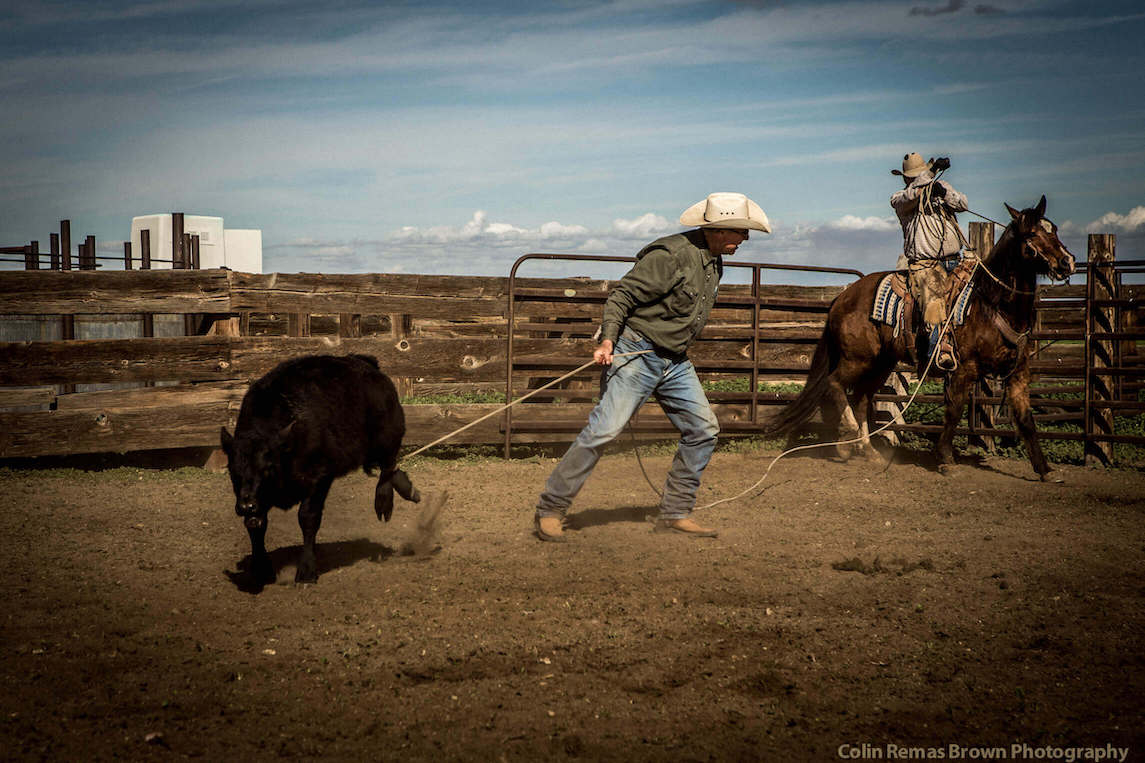}} & \multicolumn{1}{c}{\includegraphics[height=2.2cm,width=3.6cm]{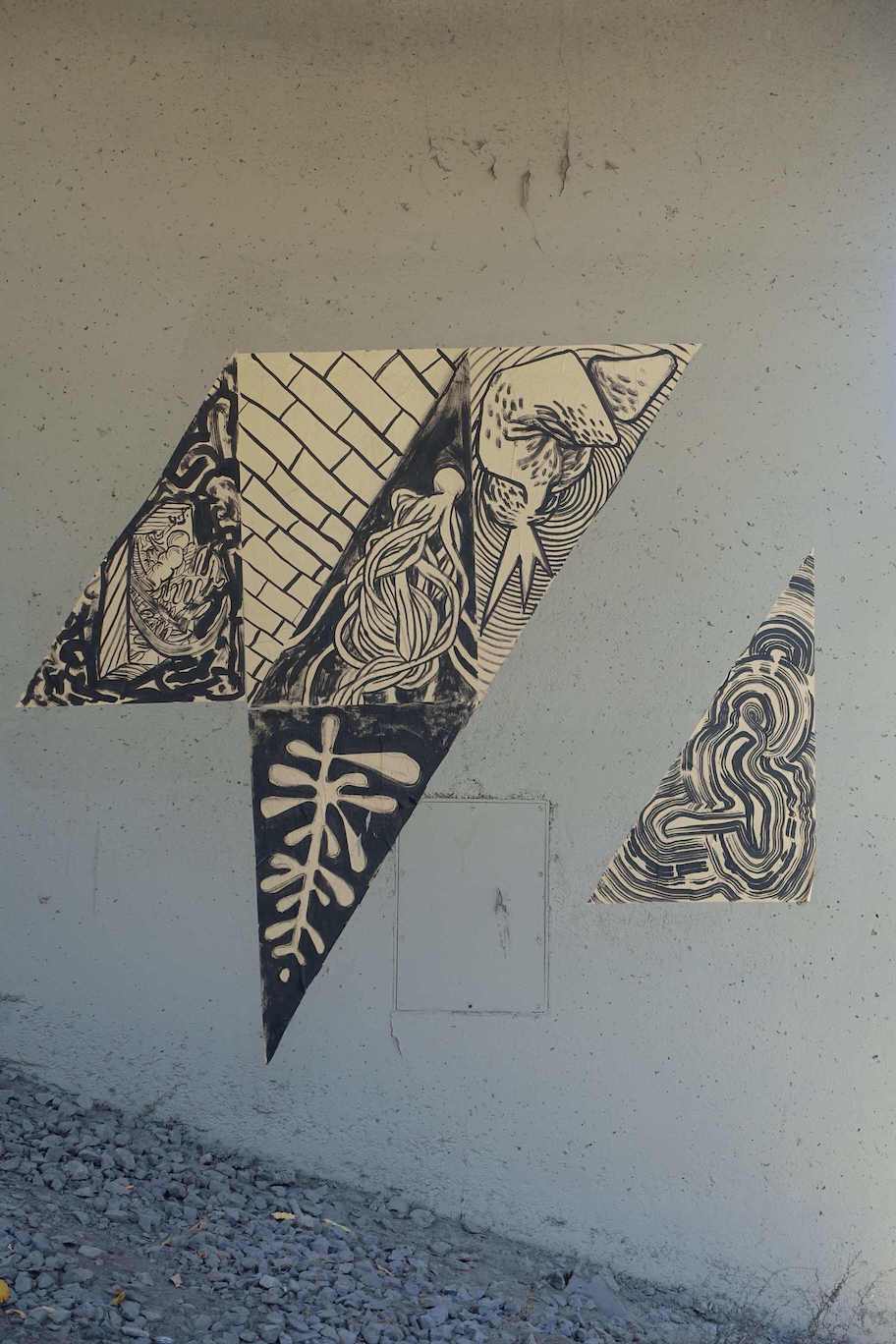}}\\
          Man sitting by his artwork looking at a large statue of a man on a horse in a royal courtyard. &  Woman with an umbrella reading a book sitting in the grass in front of a city skyline. &  Cowboy on a horse and cowboy on the ground working together to lasso a calf in a pen. &  Black and white artwork painted on a blue wall. \\
    \end{tabular}
    \caption{Examples of the ground truth captions that we collected for the \t2cc dataset. (Photo credits from left to right, top to bottom: Sharron Mollerus, Northfielder, George M. Groutas, davebloggs007, Tim Adams, Brisbane City Council, Colin Brown, Guilhem Vellut)}
    \label{fig:t2_gt_samples}
\end{figure*}

%% file: model_figure.tex
\begin{figure*}[tb]
    \small
    \centering
    \includegraphics[scale=0.48]{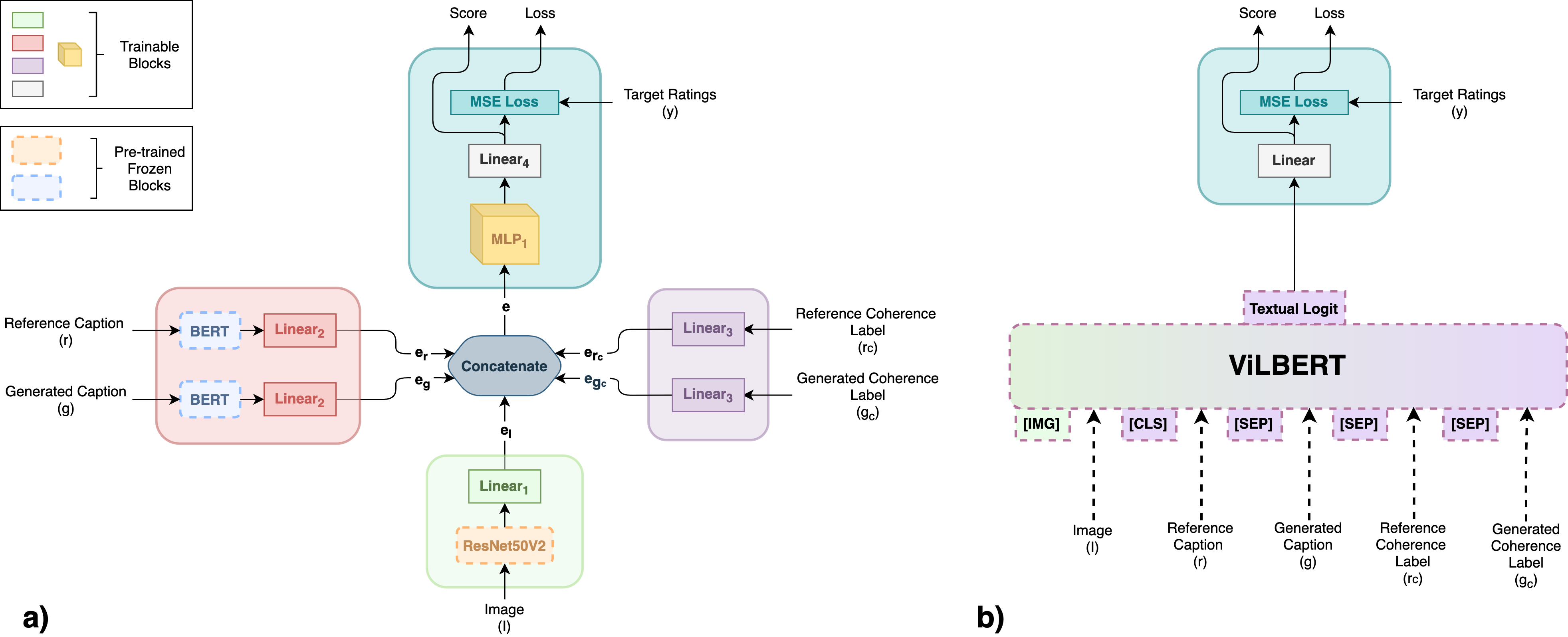}
    \caption{
      An illustration of different flavors of \ourmodel~that outputs a score for the generated caption given the image, reference caption, and the coherence-labels for both the captions.  (a) COSMic Vanilla uses only global textual and visual features, while (b) COSMic ViLBERT uses combined visio-linguistic features with both local and global focus.
      This model takes into account the information goals (determined by coherence-labels) for both the captions when comparing the generated caption to the reference for evaluation.
    }
    \label{fig:models}
\vspace{-10pt}
\end{figure*}

%% file: 03-data.tex
\section{Data Collection}
\label{sec:data}
We collect two datasets: human judgments for image captions that are generated by coherence-aware captioning systems using Conceptual Captions dataset; and ground-truth labels for the Open Images dataset. With Conceptual Captions corpora we fine-tune ViLBERT with ratings and show that addition of coherence relations can make automated scoring closer to human scoring. We use OpenImages corpora to reinforce that multimodality and coherence relations have significant contributions to scoring out-of-domain datasets, as well.



\paragraph{Protocol}
We hired two expert linguists for data annotation and designed an annotation website to facilitate the annotation procedure. They are native English speakers who identify themselves as of White and Latino ethnicity. The code \footnote{\url{https://github.com/Merterm/COSMic}} of the annotation website, and the details of the protocol is publicly available. The study has been approved by our institution's human subject board.



\paragraph{Conceptual Captions Score Annotation}
\label{sec:annotation}

We have collected ratings on the quality of different image descriptions with coherence labels for a subset of 1000 images from the Conceptual Captions (CC) training dataset \cite{ng2020understanding}. With this paper, we are publishing this dataset as a benchmark for evaluation metrics that are coherence-aware. The set-up of the data collection is as follows: CC images are input into a caption-generation model created by \newcite{alikhani-etal-2020-cross}. This model generates coherence-aware descriptions for input images in 4 different coherence classes of \texttt{Meta}, \texttt{Visible}, \texttt{Subjective}, \texttt{Story}.
%
%
%
%
These 4,000 image/caption pairs are then presented to human annotators who are asked to select the correct coherence label for each pair:\\

\begin{itemize}[noitemsep,topsep=0pt]
    \itemsep0em 
    \item \textit{Meta:} the caption talks about when, where, and how the picture is taken. \textit{Meta-talk} in \newcite{https://doi.org/10.1111/j.1475-682X.1980.tb00021.x} 
    \item \textit{Visible:} the caption is true just by looking at the picture. \textit{Restatement} relation in \newcite{prasad-etal-2008-penn}.
    \item \textit{Subjective:} the captions is the matter of opinion. \textit{Evaluation} relation in \newcite{hobbs-1985}.
    \item \textit{Story:} text and image work like story and illustration. \textit{Occasion} relation in \newcite{hobbs-1985}.
\end{itemize}

After the annotator selects a specific coherence label from the above, we ask them to rate the quality of the captions, given the label, on a scale of 1 to 5. We use these annotations as training data for our coherence-aware captioning metric, \ourmodel. We call this data we annotated \cc~(Ratings for Conceptual Caption).

To calculate the Cohen's $\kappa$ agreement measure, we selected 150 images randomly and assigned them to two annotators. The Kappa coefficient is $\kappa= 0.89$ which indicates a substantial agreement \cite{viera2005} 


\paragraph{OpenImages Ground Truth Captions}
To create an out of domain test set we asked our annotators to write \textit{Visible} captions for 1,000 images\footnote{The same subset, named T2, was used for the CVPR-2019 Workshop on Conceptual Captions, www.conceptualcaptions.com.} from the OpenImages dataset \cite{Kuznetsova_2020}. We call this dataset \t2cc (\textbf{C}orpus of \textbf{O}pen\textbf{I}mages with \textbf{N}atural descriptions). A sample of these ground truth captions written by our expert linguists are presented in Figure~\ref{fig:t2_gt_samples}. We use this dataset to test \ourmodel~and other learned metrics in Section~\ref{sec:experiments} and present our benchmark results in Table~\ref{tab:test_results}.

%% file: 04-model.tex
\section{Method}
\label{sec:models}

The goal of a coherence-aware image captioning metric is to predict a score for the generated caption given the image, reference caption, and coherence relations of one generated caption and one reference caption. This metric function $M$ can be formalized as predicting a score $s$ as follows:
\begin{equation}
\label{eq:metric}
    s = M (I, g, r, g_c, r_c; \theta)
\end{equation}
where the metric is defined by parameters $\theta$, and where
the model inputs are defined as $I$ being the image being captioned, $g$ and $r$ the generated and reference captions, respectively. $g_c$ and $r_c$ are the coherence relations for $g$, $r$ respectively.


We now describe the architecture of our coherence-aware image captioning metric, COSMic (\textbf{CO}herence-\textbf{S}ensitive \textbf{M}etric of \textbf{i}mage \textbf{c}aptions). 
It has two flavors --- a ViLBERT-based model pre-trained on large multimodal data, and a baseline Vanilla version, as illustrated in Figure~\ref{fig:models}. 
Both are trained on \cc~training data (Section~\ref{sec:data}) with normalized human annotated rating to obtain the model's target score.

\subsection{COSMic ViLBERT}
\label{sec:vilbert}
ViLBERT \cite{vilbert} is a multimodal feature learning model pre-trained on 3.3 million Conceptual Captions image and captions data.
It is trained for masked multi-modal learning and multi-modal alignment prediction and demonstrates strong performance on several downstream multimodal tasks such as VQA, VCR, grounding, and image retrieval.
For this reason we use a pre-trained ViLBERT to embed our multimodal inputs shown in Equation~\ref{eq:metric} with changes to incorporate both the captions and coherence relations.

For input image ($I$), we use the same process as ViLBERT.
We use a Faster R-CNN \cite{ren2016faster} model pre-trained on Visual Genome \cite{krishnavisualgenome} to detect objects regions and extract features.
The sequence of these image features is denoted as $I'$ with 100 bounding box features where each element is $R^{2048}$.
Similar to ViLBERT, we use the special token \textit{[IMG]} to denote the beginning of the bounding box features list.

For input captions ($g$, $r$) and coherence labels ($g_c$, $g_r$), the sequence begins with the special token \textit{[CLS]} followed by input text embeddings.
Each of our text inputs are tokenized and embedded using ViLBERT's input text pre-processing and denoted as $g'$, $r'$, $g_c'$, $g_r'$ for $g$, $r$, $g_c$ and $g_r$ respectively. 
Note that the coherence labels are processed as text inputs such as \emph{``Visible''} and \emph{``Story''} which allows the model to use its pre-trained representations of these concepts.
Each of these input sequences are separated by the special token \textit{[SEP]} to form our input sequence.



Hence, our input to ViLBERT is of form:

\begin{center}
    \small
    $v = $ (\textit{[IMG]}, $I'$, \textit{[CLS]}, $r'$, \textit{[SEP]}, $g'$, \textit{[SEP]}, $r_c'$, \textit{[SEP]}, $g_c'$)
\end{center}

We use a linear layer with sigmoid activation on ViLBERT's output text logits to compute COSMic's output metric score ($s$).
\begin{equation}
    s = \text{Linear} ( \text{ViLBERT} (v) )
\end{equation}
\noindent

During training, we fine-tune ViLBERT and the output linear layer in an end-to-end fashion by minimizing the Mean-Squared error between the output score, $s$ and the corresponding reference score, $y$, on the \cc~dataset.

\subsection{COSMic Vanilla}
The COSMic ViLBERT approach above takes advantage of multimodal pre-training on the Conceptual Captions dataset to embed the image and text inputs.
As a simpler baseline, we now present COSMic Vanilla which independently embeds the input image and text to be later combined for score computation with no end-to-end training.

To extract image features, we use a ResNet50v2 \cite{DBLP:journals/corr/HeZRS15} model pre-trained on ImageNet \cite{imagenet_cvpr09} and linearly transform the global image representation to 512-dimensional space.
\begin{equation}
    e_I = \text{Linear}_1 ( \text{AveragePool} ( \text{ResNet} (I) ) )
\end{equation}

In our textual feature extraction module, we embed $g$ and $r$ independently with a pre-trained BERT-Large-512 model.
We use the \emph{[CLS]} token embedding as 1024 dimensional caption-level representation in each case and transform them to 512-dimensional space.
\begin{equation}
\begin{aligned}
    e_g &= \text{Linear}_2 ( \text{BERT}_\text{CLS} ( g ) ) \\
    e_r &= \text{Linear}_2 ( \text{BERT}_\text{CLS} ( r ) )
\end{aligned}
\end{equation}

In our coherence label embedding module, $g_c$ and $r_c$ are each represented as one-hot vectors such that the dimensions correspond to labels \textit{Meta}, \textit{Visible}, \textit{Subjective} and \textit{Story}.
Each is embedded into a 512-dimensional space.
\begin{equation}
\begin{aligned}
    e_{g_c} &= \text{Linear}_3(g_c)  \\ e_{r_c}&=\text{Linear}_3(r_c)
\end{aligned}
\end{equation}

We thus obtain the 5 vectors (each $R^{512}$), representing one of the inputs of Equation~\ref{eq:metric}.
We concatenate and use a feed-forward network with progressively smaller hidden layers of sizes $[512, 256, 128, 64, 32, 16, 8]$, each with ReLU \cite{DBLP:journals/corr/abs-1803-08375} activation.
The output score, $s$, is computed by a final linear layer on top of the above network.

\begin{equation}
\begin{aligned}
    e &= \text{concat}([e_I, e_g, e_r, e_{g_c}, e_{r_c}])) \\
    s &= \text{Linear}_4 ( \text{MLP}_1 (e) )
\end{aligned}
\end{equation}
\noindent
where $e \in R^{2560}$ and $s \in R$.

To understand the role of each component of this implementation, we further deconstruct each module in ablation experiments described in Table~\ref{tab:ablation}.

\input{result_tables}

\subsection{Coherence-aware Captioning Systems}
In order to experiment with \ourmodel, we generate our own captions. In this section we describe the coherence-aware captioning systems used to generate these image captions for the training and testing of \ourmodel.

For our base captioning system, we use the state-of-the-art coherence-aware captioning system introduced by \cite{alikhani-etal-2020-cross}.
It uses a Transformer-based \cite{transformer} encoder-decoder architecture where the encoder inputs are (1) global image features, (2) image labels, and (3) coherence label.
The coherence-label also serves as the first input token for the decoder which generates the output captions.
We set the coherence label to the groundtruth relation at training time, and the desired relation at inference time.
We use the Conceptual Captions dataset \cite{sharma-etal-2018-conceptual} with machine-generated coherence labels for training this captioning system.
To obtain the coherence labels above, we closely follow \cite{alikhani-etal-2020-cross} to train a coherence classifier on the Clue dataset \cite{alikhani-etal-2020-cross} that provides around 4K human annotated (image, caption, relation) triplets.
We present two caption-generation systems in this section.

\paragraph{Base-systems family} 
A family of 4 captioning systems is created by setting the coherence-label to \textit{Meta}, \textit{Visible}, \textit{Subjective} or \textit{Story} in the base captioning model described above.
These are considered different captioning systems because the information content and discourse goals, as controlled by the coherence label, are different.

\paragraph{Lite-systems family}
We remove the global image features from the base model's input to obtain a smaller, light-weight (lite) model.
Similar to the base model, we obtain a family of 4 captioning systems by changing the coherence-label.

In Section~\ref{sec:experiments}, we study the order in which several image captioning metrics rank these 8 systems.
The goal is to identify the metric that agrees the most with the groundtruth rankings based on human assessments.

\subsection{COCO-trained Captioning System}
COSMic's training data, \cc, is based on Conceptual Captions and it is coherence-aware.
To test the model's generalization capability, we use a captioning system trained on MS COCO \cite{coco}.
Since \ourmodel~expects an input coherence label, and COCO captions are \textit{Visible} style by design, we set the label to \textit{Visible}.
Specifically, we use the Bottom-Up Top-Down (BUTD) Attention model \cite{Anderson_2018_CVPR}. 
This helps study how well \ourmodel~generalizes to other captioning datasets and coherence-agnostic captioning systems.

%% file: result_tables.tex
\begin{table*}[ht]
\scriptsize
\centering
\begin{tabular}{c|p{0.5cm}ccccccccccccc}
 \toprule
 \multicolumn{2}{c}{System} & \multirow{4}{0.8cm}{\centering Avg. Hum. Rating} & \multicolumn{12}{c}{Metrics} \\
 \cmidrule{1-2} \cmidrule{4-15}
 \multicolumn{1}{l|}{Model} & \multicolumn{1}{p{0.8cm}}{Coh. Label} & &  \multirow{2}{*}{B$_1$} & \multirow{2}{*}{B$_2$} & \multirow{2}{*}{M} & \multirow{2}{*}{R$_L$} & \multirow{2}{*}{C} & \multirow{2}{*}{S} & \multirow{2}{*}{BR} & \multicolumn{1}{p{0.5cm}}{BS-F} & \multicolumn{1}{p{0.8cm}}{COSMic Vanilla} & \multicolumn{1}{p{0.8cm}}{COSMic ViLBERT} & \multicolumn{1}{p{0.8cm}}{COSMic Vanilla+} & \multicolumn{1}{p{1cm}}{COSMic ViLBERT+} \\
 \midrule
BUTD & Visible & 2.191 & .163 & .077 & .049 & .160 & .092 & .030 & -.877 & .863 & .706 & .796 & .522 & .641 \\
 \midrule
 \multirow[m]{4}{*}{Base} 
 & Visible      & 3.532 & .050 & .025 & .019 & .066 & .020 & .002 & -1.114 & .862 & .696 & .777 & .516 & .614 \\
 & Meta         & 3.213 & .041 & .000 & .012 & .063 & .012 & .000 & -1.059 & .863 & .548 & .727 & .505 & .602 \\
 & Subj.         & 2.830 & .033 & .012 & .011 & .057 & .017 & .000  & -1.197 & .849 & .323 & .421 & .358 & .403 \\
 & Story        & 2.915 & .029 & .000 & .017 & .058 & .013 & .000 & -1.304 & .842 & .533 & .629 & .482 & .527 \\
 \midrule
 \multirow[m]{4}{*}{Lite} 
 & Visible      & 3.298 & .028 & .011 & .013 & .053 & .011 & .000 & -1.101 & .863 & .684 & .784 & .515 & .604 \\
 & Meta         & 2.830 & .026 & .010 & .008 & .055 & .015 & .000 & -1.084 & .859 & .548 & .748 & .511 & .565 \\
 & Subj.         & 2.298 & .039 & .012 & .019 & .066 & .024 & .003 & -1.217 & .849 & .364 & .451 & .379 & .419 \\
 & Story        & 2.426 & .036 & .000 & .018 & .062 & .021 & .000 & -1.362 & .842 & .568 & .666 & .499 & .519 \\
\midrule
 \multicolumn{2}{p{1.7cm}}{\textbf{Kendall's \ \ \ \ \ \ Correlation ($\tau$)}} & \multirow[m]{2}{*}{1.000} & \multirow[m]{2}{*}{.071} & \multirow[m]{2}{*}{.154} & \multirow[m]{2}{*}{.036} & \multirow[m]{2}{*}{-.036} & \multirow[m]{2}{*}{-.571} & \multirow[m]{2}{*}{-.052} & \multirow[m]{2}{*}{.286} & \multirow[m]{2}{*}{.445} & \multirow[m]{2}{*}{.571} & \multirow[m]{2}{*}{.546} & \multirow[m]{2}{*}{.667} & \multirow[m]{2}{*}{\textbf{.764}}\\

 \bottomrule
\end{tabular}
\caption{
    System-level scores for 9 different image captioning systems as evaluated by human annotators and various captioning metrics. 
    Bottom-Up Top-Down (BUTD) is trained on COCO, while others are trained on the Conceptual Captions (CC) dataset.
    The evaluation however is conducted on \t2cc dataset, which is out-of-domain for both COCO and CC.
    This domain shift causes the n-gram based metrics (e.g. BLEU, ROUGE, CIDEr) to assign very low scores to otherwise correct captions (See Table~\ref{tab:instances}).
    Whereas embedding based metrics (e.g. BLEURT, BERTScore and \ourmodel) do not suffer from this limitation.
    Since all metrics have different scales, instead of absolute scores, we use Kendall Rank Correlation to measure agreement with human scores.
    Model names are abbreviated as follows: B$_1$: Bleu$_1$, B$_2$: Bleu$_2$, M: METEOR, R$_L$: ROUGE$_L$, C: CIDEr, S: SPICE, BR: BLEURT, BS-F: BERTScore F1. 
    COSMic models with '+' denote application of data augmentation to remove training data bias. More metrics and detailed results can be found on the code repository.
}
\label{tab:test_results}
\vspace{-6pt}
\end{table*}

%% file: 05-experiments.tex
\section{Experiments}
\label{sec:experiments}
Here, we describe the experimental setup to compare \ourmodel~with other metrics. 
As outlined in Section \ref{sec:data} and \ref{sec:models}, we use the \cc~data to train our models, and \t2cc to test \ourmodel~and other metrics. 
We have several baseline metrics that we compare to, which can be found on Table~\ref{tab:test_results}. 
\subsection{Model Training Setup}
We implement \ourmodel---as described in Section~\ref{sec:models}---with PyTorch \cite{NEURIPS2019_9015} and train on a GTX1080 GPU.
We pre-compute BERT\footnote{\url{https://github.com/google-research/bert}} and ResNet\footnote{\url{https://www.tensorflow.org/api_docs/python/tf/keras/applications/ResNet50V2}} features using their TensorFlow \cite{tensorflow2015-whitepaper} implementations. We use the public ViLBERT\footnote{\url{https://github.com/facebookresearch/vilbert-multi-task}} implementation. We use a batch size of 4, and a learning rate of $2\times10^{-6}$ for fine-tuning ViLBERT and use RAdam optimizer and stop the training when the validation score does not change for 3 epochs.
For COSMic Vanilla, we train with a batch-size of 10, Adam optimizer \cite{kingma2017adam} with a base learning rate of $10^{-3}$ that decays by a factor of $10^{-2}$ every 10 epochs.
We observe that the Vanilla converges in approximately 100 epochs and ViLBERT converges in 9 epochs. ViLBERT has 250 million parameters. COSMic Vanilla includes 3,062,913 trainable parameters. Pre-trained BERT-Large and ResNet50V2 have an additional 350 million parameters. 
The setup for coherence-aware captioning models to obtain machine-generated captions for our study is the same as \cite{alikhani-etal-2020-cross}.




\subsection{Baseline Captioning Metrics}
To benchmark \ourmodel, we compare it with other learned metrics. In this section we describe these various metrics traditionally used for measuring image captioning systems. None of these metrics were designed to support the coherence relations of the reference or generated captions. These serve as baselines for \ourmodel.

\paragraph{N-gram based}
The most popular image captioning metrics are based on precision and recall of n-grams from generated and reference captions.
We compare with Bleu$_1$, Bleu$_2$, Bleu$_3$, Bleu$_4$ \cite{guo-hu-2019-meteor}, ROUGE$_L$ \cite{lin-2004-rouge}, CIDEr \cite{vedantam2015cider}, and SPICE \cite{anderson2016spice}.
We compute these using their popular open-source implementation\footnote{\url{https://github.com/tylin/coco-caption}}.

\paragraph{BLEURT} 
We use a pre-trained BLEURT model\footnote{\url{https://github.com/google-research/bleurt}} as a baseline for our work.
Unlike N-gram based approaches, BLEURT uses BERT-based word embeddings which are robust to variations in surface word realizations between the reference and generated captions.
We do not do any fine-tuning for this baseline.


\paragraph{BERTScore} 
BERTScore\footnote{\url{https://github.com/Tiiiger/bert_score}} uses a pre-trained BERT model to embed the reference and generated captions.
Text-level similarity scores are then computed by matching the tokens' output embeddings.

Please note that for both BERT-based baselines above (BLEURT, BERTScore), we use the BERT-Large-512 size model.

\input{t2_comparison}

\subsection{\t2cc-based Evaluation Setup}
We use each baseline metric and \ourmodel~to score the 8 different image captioning systems described in Section~\ref{sec:models} on the same set of test images with reference captions.
Note that the range and scale of each metric is different, however they are all monotonously increasing functions of model quality.
So in our study, we do not analyze the absolute score assigned by these metrics, but only their ranks. We also ask human annotators to rank these 8 captioning systems on the same set of test images. The ranks assigned by a higher performing metric will align better with the ranks from human annotators.

Since the captioning systems above are trained on Conceptual Captions or COCO, we use image/caption pairs from \t2cc for an out-of-domain evaluation.
A subset of 50 random images is used to rank the captioning systems as described above, resulting in 400 machine-generated captions total for the 8 captioning systems.
These were then evaluated by human annotators using the process described in Section~\ref{sec:data}.
The human-scored system level performance for each captioning system on this test set is reported in Table~\ref{tab:test_results} in ``Average Human Rating''.

We measure the alignment between metric-assigned and human-assigned scores using the Kendall \cite{kendall} correlation coefficient. In order to calculate the score, we first aggregate all the sample scores and average them. Then we calculate the Kendall tau score using the SciPy 1.7.1 implementation. The score is calculated between two vectors, first of which is the average human ratings for 8 models and the second being the investigated metric scores for 8 models in the following order:$[$Base$_{Visible}$, Base$_{Meta}$, Base$_{Subjective}$, Base$_{Story}$, Lite$_{Visible}$, Lite$_{Meta}$, Lite$_{Subjective}$, Lite$_{Story}]$. Due to the small sample size, Kendall correlation is the most suitable correlation measure.

A key measure of the success of an automatic evaluation metric is whether it makes the same decision about which system is better in a head-to-head evaluation as we would get from a human-subjects evaluation.  If each system is evaluated based on its average score, then success comes when the average computed metric correlates closely with the average human-ranking.  In particular, we measure the alignment between metric assigned and human assigned scores using the Kendall score, following the work of \cite{sellam-etal-2020-bleurt}. 

%% file: t2_comparison.tex
\begin{figure*}[b]
    \small
    \centering
    \begin{tabular}{cp{3cm}p{3cm}p{3.5cm}p{3cm}}
         & \centering{\includegraphics[width=3cm,height=2.4cm]{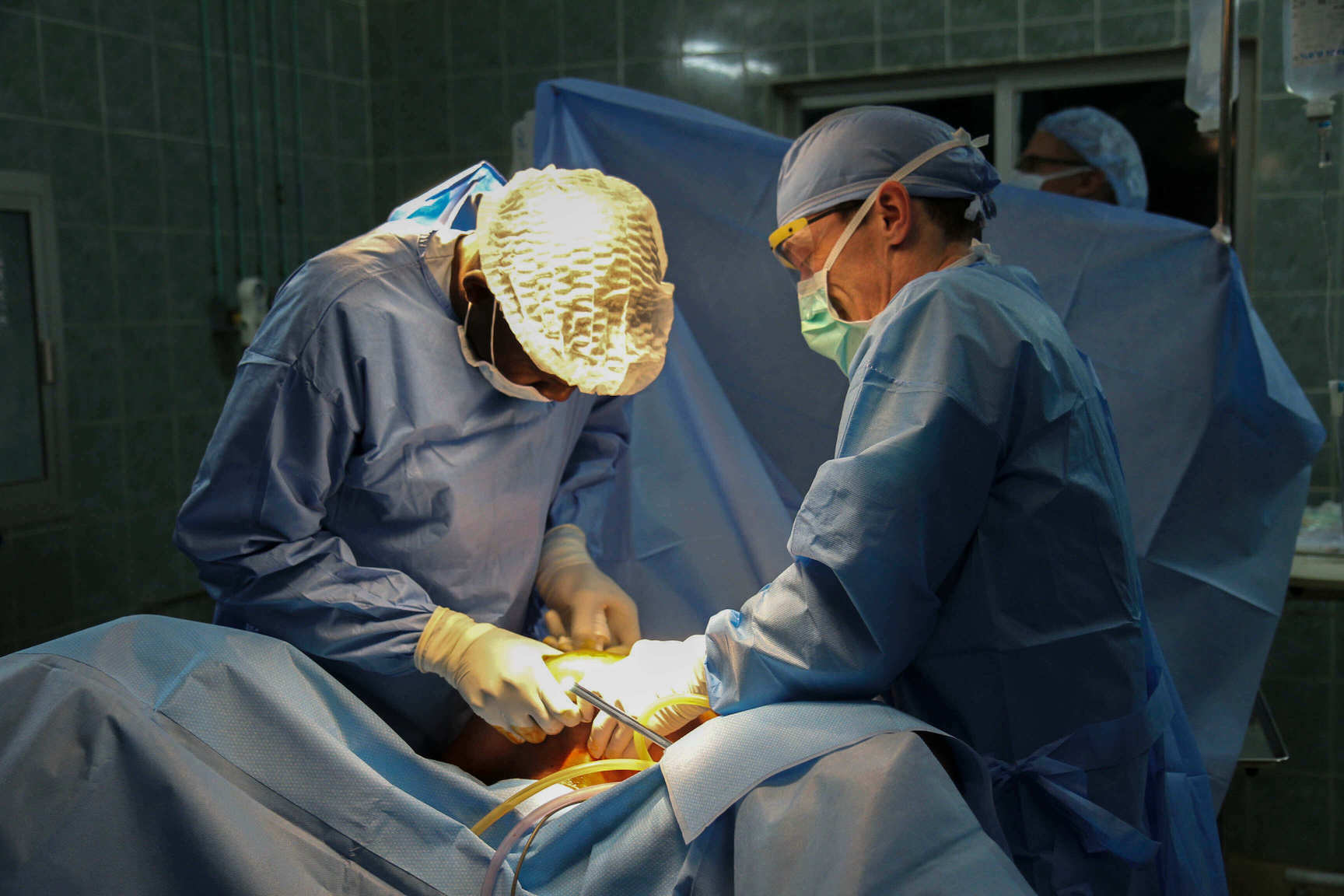}} & 
         \centering{\includegraphics[width=3cm,height=2.4cm]{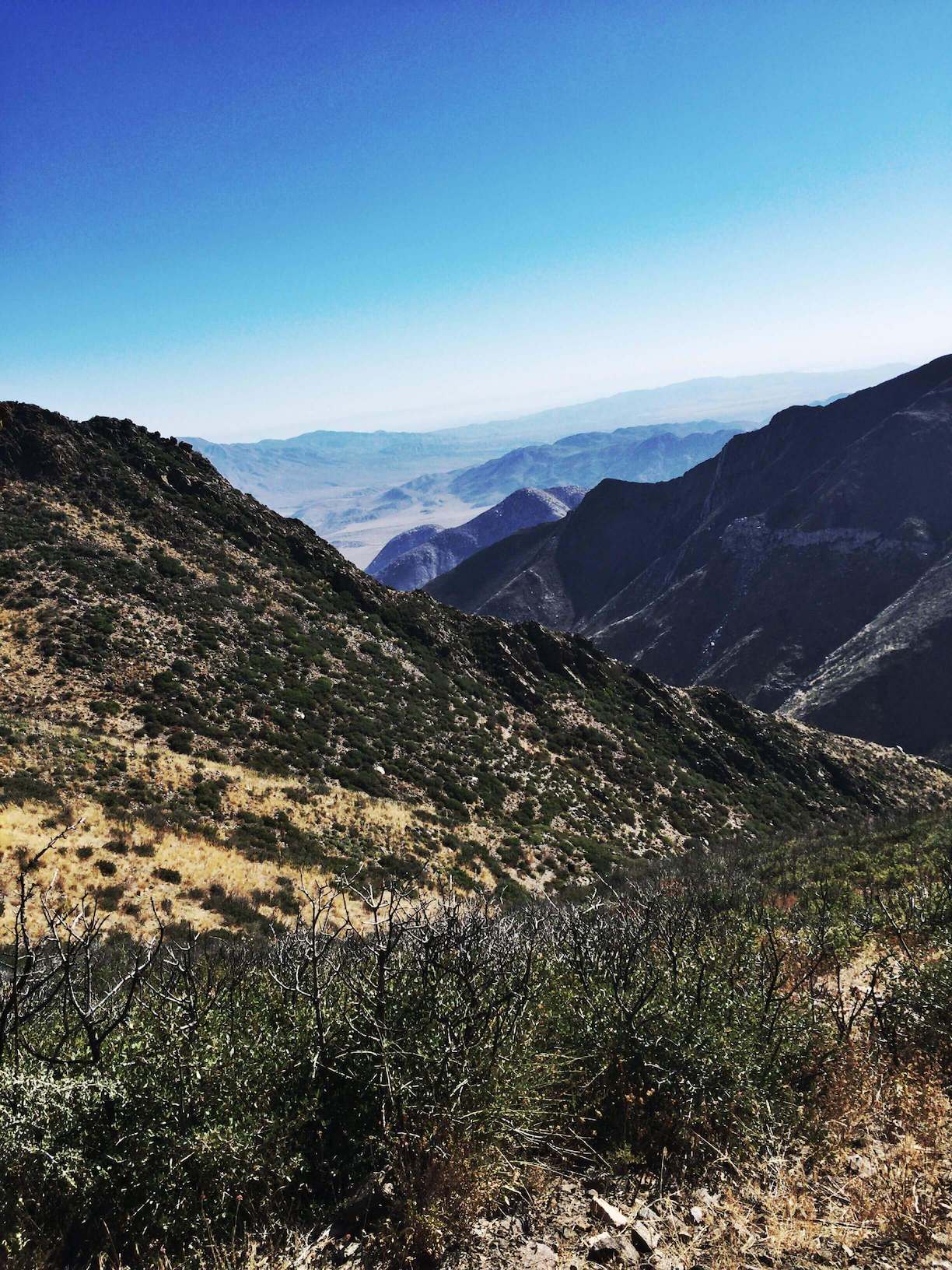}} & 
         \centering{\includegraphics[width=3cm,height=2.4cm]{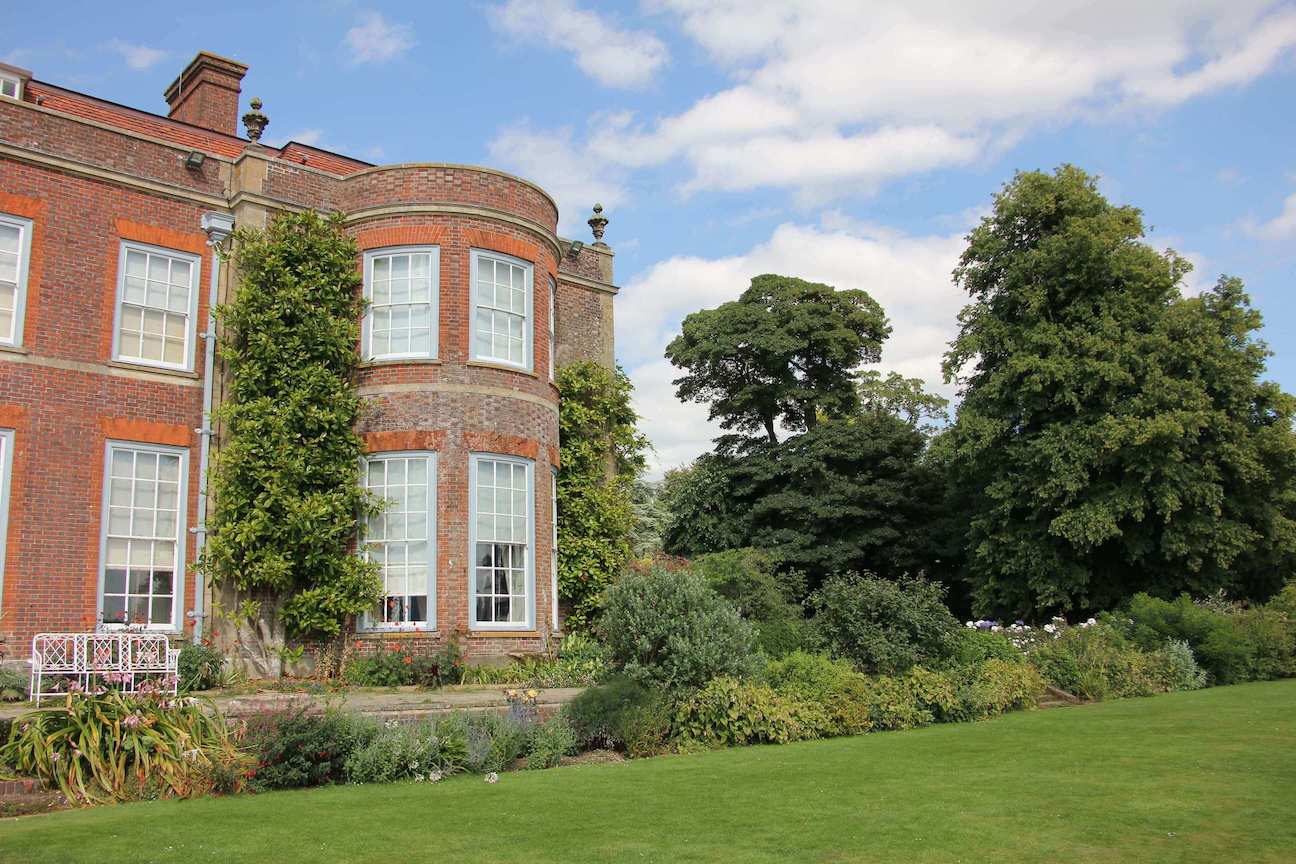}} & 
         \includegraphics[width=3cm,height=2.4cm]{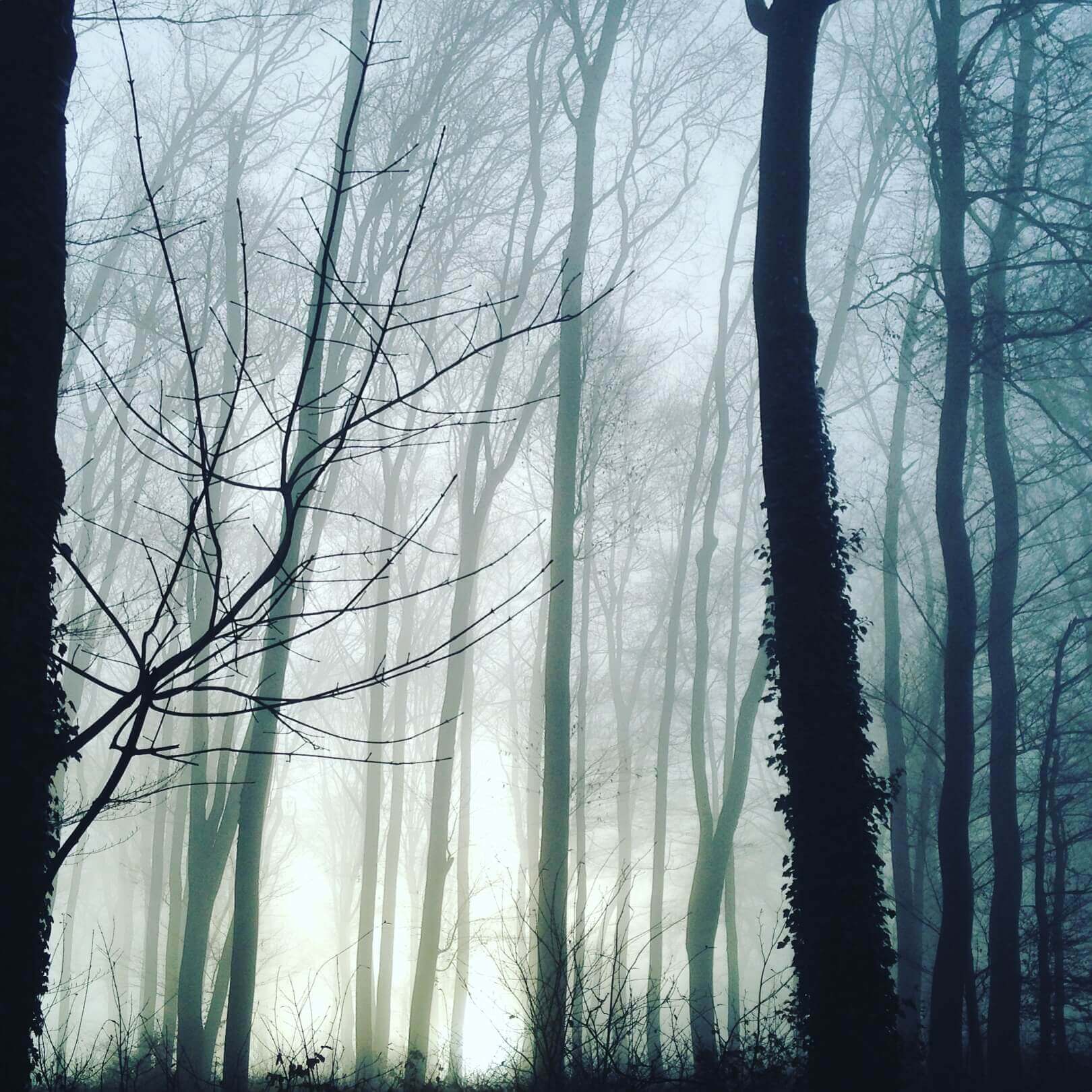} \\
         \textbf{Reference} & two men in scrubs performing surgery. & mountains in front of a clear blue sky. & large brick building next to a green lawn and big trees. & a foggy forest. \\
         \textbf{Generated} & surgeons operating on a patient. & mountain range as seen from the trail.  & the front of the house. & light shining through the trees. \\
    \end{tabular}
    \caption{
        Illustration of \t2cc reference captions and corresponding outputs of the Base-Visible model.
        Though the generated captions are correct, an n-gram based metric such as CIDEr assigns them a very low score due to the variations in surface word realizations.
        See Table~\ref{tab:test_results} for average scores over the test set. (Photo credits, from left to right: U.S. Army Africa, Gabriel, Fr James Bradley, Rosmarie Voegtli)}
    \label{tab:instances}
\end{figure*}

%% file: 06-results.tex
\section{Results}
\label{sec:results}
Table~\ref{tab:test_results} presents the results of the \t2cc-based study.
The last row reports the Kendall correlation coefficient between the scores assigned by the metric and humans.

All N-gram based metrics, such as BLEU and CIDEr, fail to adapt to the out-of-domain ground-truth captions from \t2cc.
This results in a relatively flat distribution of system-level scores concentrated close to 0, and hence low correlation coefficients.
CIDEr has a highly negative Kendall's $\tau$, which denotes a strong negative association with human judgements. This is partly due to low ($\sim$0.01) and hence noisy CIDEr scores. (Figure~\ref{tab:instances} provides example cases that illustrate this argument.) 


Embedding-based methods, BLEURT and BERTScore, do not suffer from this limitation resulting in more meaningful scoring of systems and hence higher correlation with human scores.
However, by design, both these metrics are agnostic to coherence-labels and the input image.
\ourmodel, which is coherence-aware, obtains the highest correlation with human scores.
COSMic ViLBERT has the highest Kendall's correlation among all of our models. 
COSMic Vanilla performs the second best among our models and it performs better than the rest of the models in terms of Kendall's correlation.


\paragraph{Data Augmentation}
The raw \cc~training data has a coherence-level bias as demonstrated by the average COSMic score for each class --- \textit{Visible} (0.622), \textit{Meta} (0.459), \textit{Subjective} (0.236) and \textit{Story} (0.397).
This reflects the human annotators' bias towards liking \textit{Visible} captions the most, and \textit{Subjective} captions the least, which is expected.
However, training \ourmodel~on this data injects the same coherence-bias into the model which is undesirable.
As presented in Table~\ref{tab:test_results}, both flavors of COSMic (without the `+') assign high scores to \textit{Visible} captioning systems.

To mitigate this issue, we algorithmically augment the training data to bring the average scores for each coherence class to comparable values.
We achieve this by pairing images with random captions from the coherence class and assigning them a score of 0.
This is a valid training sample because the randomly sampled caption does not describe the said image and serves as a negative sample.
With these operations, the class bias is significantly reduced --- \textit{Visible} (0.459), \textit{Meta} (0.439), \textit{Subjective} (0.328) and \textit{Story} (0.425).
The \ourmodel~columns in Table~\ref{tab:test_results} with `+' denote that this data augmentation approach improves ranking of captioning systems leading to better alignment with human judgements.

\paragraph{Ablation Study}
Table~\ref{tab:ablation} reports the performance of \ourmodel~Vanilla without coherence-labels and/or the image as model inputs.
We find that removal of image features affects \ourmodel's performance, showing the important contribution of images.
The performance deteriorates significantly when the coherence-labels are removed from the model ("No $r_c, g_c$" column in Table~\ref{tab:ablation}).
This demonstrates that \ourmodel~ successfully integrates coherence-relations in the caption scoring process.

\begin{table}
    \small
    \centering
    \begin{tabular}{c|p{0.8cm}|ccccc}
 \toprule
 \multicolumn{2}{c|}{System} & \multicolumn{3}{c}{\ourmodel} \\
 \midrule
 \multicolumn{1}{c|}{Model} & \multicolumn{1}{p{0.8cm}|}{Coh. Label} & Full & No $I$ & No $c$ & No $I$ \& $c$ \\
 \midrule
 \multirow[m]{4}{*}{Base} 
 & Visible      & .516 & .447 & .434 & .442 \\
 & Meta         & .505 & .439 & .442 & .453 \\
 & Subj.        & .356 & .347 & .438 & .453 \\
 & Story        & .505 & .433 & .436 & .445 \\
 \midrule
 \multirow[m]{4}{*}{Lite} 
 & Visible      & .515 & .444 & .434 & .433 \\
 & Meta         & .511 & .434 & .447 & .464\\
 & Subj.        & .379 & .367 & .440 & .459 \\
 & Story        & .499 & .440 & .433 & .442 \\
\midrule
 \multicolumn{2}{p{1.2cm}}{\textbf{Kendall's Corr. ($\tau$)}} & \multirow[m]{2}{*}{\textbf{.667}} & \multirow[m]{2}{*}{.546} & \multirow[m]{2}{*}{-.222} & \multirow[m]{2}{*}{-.415} \\
 
 \bottomrule
\end{tabular}
    \caption{Ablation experiment results. "No $I$" represents "\ourmodel~Vanilla without image features", "No $r_c,g_c$" represents "\ourmodel~Vanilla without coherence label embeddings", finally "No $I$ \& No $r_c,g_c$" represents "\ourmodel~Vanilla without coherence label embeddings and without image features".}
    \label{tab:ablation}
\end{table}

%% file: 07-conclusion.tex

\section{Conclusion}

Our work is the first step towards designing generation metrics that respect the information goal of the generated text. We observe that a small set of examples annotated with coherence relations can provide what is needed for learning a discourse-aware generation metric. Our findings have implications for designing context-aware multimodal metrics with criteria that are closer to human ratings for evaluating machine-generated multimodal content. 

We have called attention to the challenge of learning robust generation metrics that can evaluate the output of the generation models considering the information goals. Our findings suggest that fine-tuning ViLBERT---originally trained with millions of images---with a smaller sample of coherence relations and expert-annotated scoring, automated metrics can score generated captions closer to a human rating.
The presented dataset provides the opportunity for future research in the area of image description generation, designing discourse-aware metrics, and multimodal content evaluation. 
We hope that coherence-aware text generation metrics could be used for learning better generation models (such as abstractive summarization or story generation) and could be deployed directly in machine learning pipelines to help in optimizing hyper-parameters. Ultimately, it is intended to have a generalizable model that can use a labeling mechanism---not restricted to coherence labels--- to improve applicability of generation metrics in different tasks.



\section{Ethics}
This paper describes a research prototype.  We do not work with sensitive or personal data. Our protocol was approved by our ethics board.  Human subjects participated voluntarily, undertook minimal risk, and were compensated fairly for their time.  The dataset we produced is fully anonymized.  Subjects consented to the distribution of their data as part of their participation in the research.
Technologists should think carefully before deploying our ideas in production.  Our work depends on pretrained models such as word and image embeddings.  These models are known to reproduce and even magnify societal bias present in training data.  Moreover, like many ML NLP methods, our methods are likely to perform better for content that is better represented in training, leading to further bias against marginalized groups.  We can hope that general methods to mitigate harms from ML bias can address these issues.

A distinctive complication of our work is the fact that many image--text presentations involve writers expressing subjective opinions.  By its nature, our evaluation metric assesses such subjective texts based on averages and trends across many users, which may be problematic.  Although such judgments are ultimately matters of personal taste, they are nevertheless often grounds by which hierarchies of differences are culturally encoded and enforced. Thus, a deployed subjective-caption generation system could well be unfair to users, especially if those users are not confident in their own taste or critical towards the system’s responses.  Our evaluation metric is not sensitive to such harms.

\section*{Acknowledgements}
The authors affiliated with Rutgers University were partly supported by
NSF Award CCF-19349243. Thanks to \href{https:
//www.cyber.pitt.edu/}{Pitt Cyber} for supporting this
project and the authors from the University of Pittsburgh. We also acknowledge the
Center for Research Computing at the University
of Pittsburgh for providing the required computational resources for carrying out experiments at the University of Pittsburgh.